\documentclass[a4paper]{article}
\usepackage{INTERSPEECH2021}
\usepackage{cite}
\usepackage{svg}
\usepackage{multirow}
\usepackage{url}
\title{BBS-KWS: The Mandarin Keyword Spotting System Won the Video Keyword Wakeup Challenge\thanks{* Equal contributions to this work}} 
\name{Yuting Yang$^{*}$, Binbin Du$^{*}$, Yingxin Zhang$^{*}$, Wenxuan Wang$^{*}$, Yuke Li}
\address{ Yidun AI Lab, Netease Hangzhou 310052}
\email{$\left\{yangyuting04, dubinbin, zhangyingxin03, wangwenxuan, liyuke\right\}$@corp.netease.com}

\begin{document}

\maketitle

\begin{abstract}
This paper introduces the system submitted by the Yidun NISP team to the \textit{video keyword wakeup} (VKW) challenge \footnote{\url{https://www.datatang.com/VMR}}.
We propose a mandarin \textbf{k}ey\textbf{w}ord \textbf{s}potting system (KWS) with several novel and effective improvements, including a big backbone (B) model, a keyword biasing (B) mechanism and the introduction of syllable modeling units (S). By considering this, we term the total system BBS-KWS as an abbreviation. The BBS-KWS system consists of an end-to-end automatic speech recognition (ASR) module and a KWS module. The ASR module converts speech features to text representations, which applies a big backbone network to the acoustic model and takes syllable modeling units into consideration as well. In addition, the keyword biasing mechanism is used to improve the recall rate of keywords in the ASR inference stage. The KWS module applies multiple criteria to determine the absence or presence of the keywords, such as multi-stage matching, fuzzy matching, and connectionist temporal classification (CTC) prefix score. To further improve our system, we conduct semi-supervised learning on the CN-Celeb \cite{fan2020cn} dataset for better generalization. In the VKW task, the BBS-KWS system achieves significant gains over the baseline and won the first place in two tracks.
\end{abstract}

\section{Introduction}

Keyword spotting is a task to detect predefined words from continuous speech. It is of importance for human-computer interaction, and is widely used in various smart devices and voice retrieval systems. In recent years, due to the rapid development of artificial intelligence technology, keyword spotting has also achieved promising developments in many scenarios, such as smart speaker, voice assistant, $etc$.

In the past few decades, researchers have proposed kinds of techniques to improve the performance of keyword spotting systems. The first is the family of Query-by-Example (QbyE) methods \cite{vasudev2015query} \cite{2017Query} \cite{zhang2009unsupervised} \cite{barakat2012improved}, which utilize the keywords speech samples to get a set of feature templates. As in the detection phase, a feature template is extracted from the test speech sample and matched with the keyword feature template via the pair templates similarity. If the similarity exceeds the threshold, it is considered as a hit. The second is the large vocabulary continuous speech recognition (LVCSR) \cite{weintraub1995lvcsr} \cite{chen2013quantifying} \cite{chen2013using} based methods, which is widely used in audio retrieval tasks. These methods transcribe speeches to texts and indexes for keywords information. In order to improve the recall rate, some improved methods \cite{thambiratnam2005dynamic} \cite{cardillo2002phonetic} introduce lattice to save multiple decoded sequences as well as position information.

In the VKW challenge, the organizer provides 1505 hours training data and 15 hours fine-tuning data. The F1 score and the actual term-weighted value (ATWV) are used to evaluate system performance. The challenge mainly focus on evaluating the accuracy and the recall rate of the keyword spotting system. In this case, we choose the LVCSR based method for this challenge.

We introduce a BBS-KWS system, which aims to improve the accuracy and the recall rate of keyword spotting in the entire system. Specifically, the equipped ASR module transcribes input speech feature to text representations, and the KWS module uses both the posterior probability of the ASR module and the transcribed text to query keyword candidates paired with the corresponding scores. Since the acoustic model has taken the custom Chinese characters into consideration, it is unnecessary to retrain the model for different keywords. Our contributions are summarized as follows:

(1) We introduce a big backbone network and syllable modeling units to improve the cross-domain performance.

(2) We introduce a keyword biasing mechanism to improve the recall rate of keywords in the speech recognition stage.

(3) In the KWS module, we utilize a series of methods to improve the recall rate of the keyword detection, including multi-stage matching and fuzzy matching.

Inspired by the references \cite{park2020improved} \cite{xie2020self}, we use semi-supervised learning on the CN-Celeb dataset \cite{fan2020cn} to solve the problem of data sparsity and the domain mismatch between the training set and the validation set. Specifically, we first train a model on the labeled data and then use it as the initial teacher model. We generate pseudo-labels for the unlabeled CN-Celeb dataset via a teacher model, and then use synthesized data to fine-tune the model. This step will be repeated for multiple rounds, and the performance of the new model on the validation set can be greatly improved.

This paper is organized as follows: In Section 2, we describe the structure of the overall BBS-KWS system, including the ASR module and the KWS module. In Section 3, we introduce the experimental setup, the semi-supervised learning and the experimental results on the VKW challenge. 

\begin{figure}[t]
  \centering
  \includegraphics[width=\linewidth]{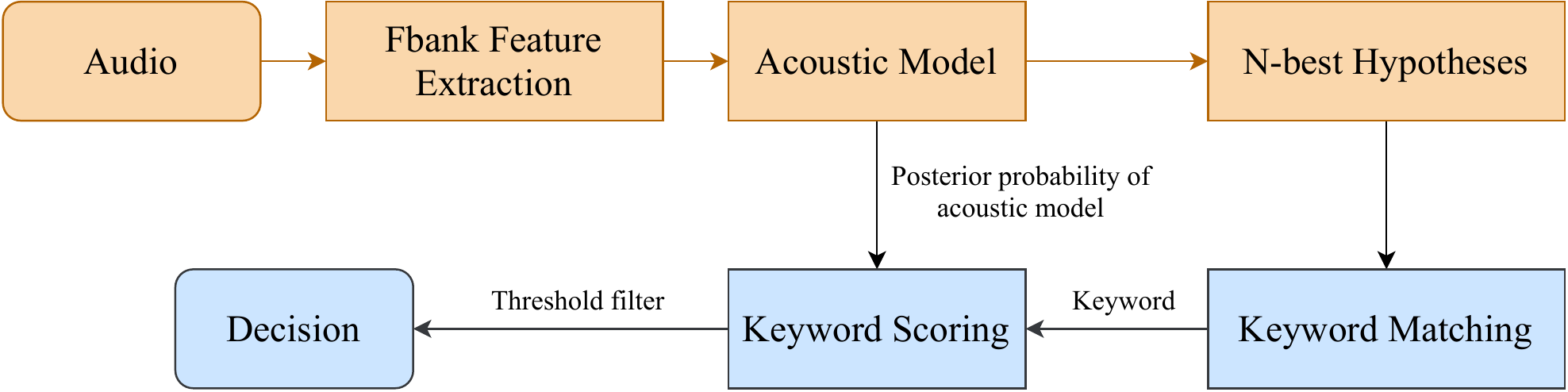}
  \caption{The pipeline of BBS-KWS system.}
  \label{fig:BBS}
\end{figure}

\section{BBS-KWS Model}

The BBS-KWS system consists of a ASR module and a KWS module, the schematic diagram of the system is provided in Figure~\ref{fig:BBS}. In the whole keyword spotting process, the ASR module converts the speech features into N-best hypotheses, and the KWS module continuously match each candidate keywords on the N-best candidate sequences. Once a keyword hits, it uses the results of the ASR acoustic model to calculate the CTC forward score as the confidence score.

\subsection{ASR Module}

The acoustic model converts the input audio features into the probability distribution over the modeling units, which is the most important part in the entire system. In our system, the N-best candidate sequences are generated by the acoustic model.

\begin{equation}
  loss = {\lambda} loss_{ctc} + (1-{\lambda})loss_{att}
  \label{eq1}
\end{equation}

The acoustic model is based on a hybrid CTC/Attention structure \cite{graves2006connectionist} \cite{kim2017joint} \cite{watanabe2017hybrid}. The input of the model is 80-dimensional filterbank features. The output of the encoder is used to calculate the CTC objective, and the output of the decoder together with the ground-truth label are utilized to obtain the CE loss. During training, the loss functions of the two branches will be linearly combined in a certain proportion. As shown in Eq.(\ref{eq1}), where $\lambda$ denotes the weight of different loss.

Different from the training process, the inference phase of BBS-KWS only uses the branch of CTC module to generate N-best hypotheses. It uses prefix beam search to generate N-best hypotheses, and uses a 4-gram language model (LM) for shallow fusion, and finally obtain N-best sequences.

\subsubsection{Big backbone}

In recent years, it is common to use the deeper models to improve the performance. For example, researchers have investigated the Transformer \cite{vaswani2017attention} \cite{dong2018speech} \cite{mohamed2019transformers} model in the field of speech recognition. They usually adopt several layers of CNN plus multiple stacked transformer sublayers. However, it is hard to further improve the model accuracy with increasing the depth of the network. Therefore, the depth of the transformer is generally in the range of 6-24, which is less than the number of network layers in the image field.

To achieve a trade-off between the accuracy and the training efficiency, the BBS-KWS system does not extend the depth of the model, but expand the width of the conformer network \cite{gulati2020conformer}. Specifically, the encoder adopt the conformer structure with a depth of 12, while the decoder has 6 identical layers. The attention dimension is increased from 256 to 512, and the number of attention heads is also increased from 4 to 8. The width of the feed-forward layer is 2048. 

\subsubsection{Keyword biasing}
Due to the factors such as context information and the frequency of tokens appearing in the training set, the rarely-used words are usually under-estimated and thus leading to the absence of them in N-best paths. This makes keyword spotting a challenging task as the rarely-used words can't pass into the final keyword determining process. To solve this issue, we utilize keyword biasing technology to perform keyword matching in real time on the process of generating N-best hypotheses. If a certain keyword is matched, the corresponding candidate sentence will be awarded with a certain score.

\begin{equation}
  W(keyword) = -{\alpha}LM(keyword) + {\beta}
  \label{eq2}
\end{equation}

We introduce a language model and use the predicted scores to adaptively assign weights to words. The N-gram language model counts the word frequency in the training data. By taking a negative number for the score of the language model, the low-frequency keyword will be assigned with high weight. Especially for long keywords, it will be segmented into several parts before calculating weights, thus avoiding the situation that some keywords cannot be matched because of their excessive lengths. As shown in Eq.(\ref{eq2}), where $LM$ is the N-gram language model, $\alpha$ and $\beta$ are parameters of the affine transformation.

\subsubsection{Syllable modeling units}

Another problem in the VKW task is that there are less in-domain data matching the test set. It is extremely difficult to fine-tune the effect of the large model on such a small amount of data. To solve this problem, we follow \cite{zhang2019investigation} \cite{2019Char} \cite{zou2018comparable} to introduce the smaller syllable modeling units than character units. At the same time, in order to maintain the accuracy at the expected character level, the character-level modeling unit is also retained.

\begin{figure}[t]
  \centering
  \includegraphics[width=\linewidth]{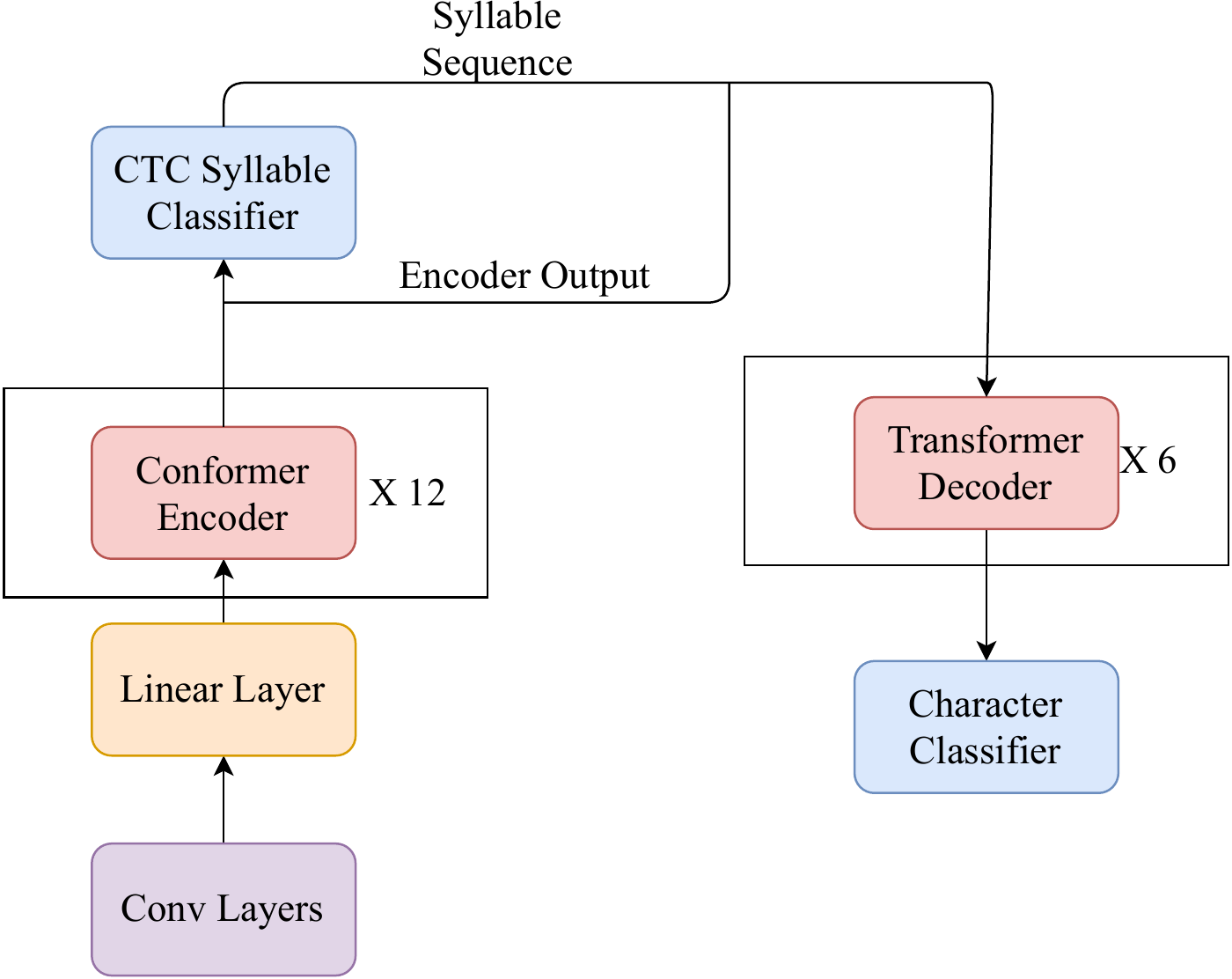}
  \caption{Hybrid syllable-character model.}
  \label{fig:p2}
\end{figure}

The CTC classifier considers syllables as modeling units, and the decoder deal with characters units. Therefore, the entire loss function is a linear combination between the syllable-based CTC loss and the character-based CE loss. The specific structure is shown in Figure~\ref{fig:p2}.

\subsection{KWS Module}

The KWS module aims to detect whether the keywords exist in the speech according to the results of the ASR module, which can be divided into two steps: matching and scoring. In the matching stage, the BBS-KWS system uses fuzzy matching and multi-stage matching to improve the recall rate of the keywords. And in the scoring stage, the BBS-KWS system uses the CTC algorithm to calculate the confidence score of the keywords. 

\subsubsection{Keyword matching and scoring}

\begin{table*}[th]
  \caption{Comparison of methods}
  \label{tab:example}
  \centering
  \begin{tabular}{|l|c|c|c|c|c|c|c|}
  \hline
		\multirow{2}{*}{Methods} & \multicolumn{3}{|c|}{F1} & \multicolumn{3}{|c|}{ATWV} \\
		\cline{2-7}
		~ & lgv & liv & stv & lgv & liv & stv \\
    \hline
    Chain model baseline&0.6781&0.6565&0.7006 &0.5171&0.6027&0.5644\\
    \hline 
    Conformer with character modeling unit&0.7646&0.7899&0.8154 &0.5534&0.6479&0.6222\\
    + LM                          &0.7988&0.8249&0.839 &0.6191&0.7183&0.6997\\
    + Length normalization&0.8236&0.849&0.8579 &0.6703&0.7784&0.7711\\
    + N-best matching&0.8247&0.8476&0.8531 &0.7019&0.8066&0.7933\\
    + SSL round 1&0.8543&0.886&0.8969  &0.7192&0.8271&0.8129\\
    + keyword biasing&0.857&0.8813&0.8898 &0.7587&0.8457&0.8356\\
    + SSL round 4&0.8644&0.8897&0.8893 &0.7677&0.8538&0.8335\\
    + Fuzzy matching&0.8681&0.8937&0.8987 &0.8078&0.8693&0.8579\\
    \textbf{+ Syllable modeling units} &\textbf{0.8886}&\textbf{0.9059}&\textbf{0.91 }&\textbf{0.82}&\textbf{0.8809}&\textbf{0.8683}\\
    \hline
    \textbf{Ensemble} &\textbf{0.8839}&\textbf{0.8973}&\textbf{0.8965} &\textbf{0.8495}&\textbf{0.9024}&\textbf{0.8895}\\
    \hline 
  \end{tabular}
\end{table*}

The KWS module performs to query keywords from the N-best results decoded by the acoustic model. If the keyword appears more than once in the N-best sequences, it is considered as a hit and passed into the next step of scoring.

\textbf{Keyword Matching:} During the matching process, BBS-KWS introduces a multi-stage matching strategy and a fuzzy matching strategy to improve the recall rate. Multi-stage matching uses the model with hybrid syllable and character modeling units for syllable matching and character matching. Keywords will be matched in both the syllable N-best sequences and the character N-best sequences, respectively. At the same time, fuzzy matching is used to improve the recall rate of keyword spotting. The dimsim \footnote{\url{https://github.com/System-T/DimSim.git}} library is used to calculate the pronunciation similarity between the decoded vocabulary and the keywords. If the distance is less than a certain threshold, the keyword is also considered as a hit.

\textbf{Keyword Scoring:} Once the keyword hits, it will be passed to the consequent scoring stage. The CTC classifier outputs the probability distribution of each frame, and uses the CTC peak information to obtain the position offset of the keyword in the speech as well as computes the probability of the keyword path. 

\begin{equation}
  S(kw) = {\sum}_{\pi:\beta(\pi)=kw}\rho(\pi) 
  \label{eq3}
\end{equation}

As shown in Eq.(\ref{eq3}), where \emph{kw} stands for keywords, $\beta$ is the CTC path compression algorithm, $\pi$ is the keyword's state path of the CTC prefix beam search, and $\rho( )$ calculate the path score. In particular, if multi-stage matching is used, once the syllable keyword and the character keyword hit at the same time, the result with a higher score will be reserved. 

The BBS-KWS applies the length normalization to the confidence score.

\begin{equation}
  Score(kw) = {S}(kw) / length(kw) 
  \label{eq4}
\end{equation}

\section{Experiments And Results}
\subsection{Experimental Setup}
The training data consists of the following three parts: (1) 1505 hours of training set (2) 15 hours of fine-tune dataset collected from Long video, short video, and live broadcast (3) 560h unlabeled data randomly selected from CN-Celeb dataset for semi-supervised learning. We use the SpecAug method \cite{park2019specaugment} during training process and speed augmentation with coefficient {(0.9, 1.0, 1.1)} for training data respectively.

The text corpus for language model consists of two parts: (1) labeled data provided by the contestant. (2) the text corpus are collected from the public website, including Wikipedia, Weibo, Douban, and Netease News. The language model is trained on 40M sentences. Both the training data and the text corpus used in restricted and unrestricted tracks are the same.

The BBS-KWS adopt the conformer structure, the attention dimension is 512, and the number of attention heads are 8. We used a 12-layer encoder and a 6-layer decoder. The $\lambda$ in Eq.(\ref{eq1}) is 0.9. During decoding, ($\alpha$,${\beta}$) in Eq.(\ref{eq2}) is (1,4), the beam size is 10. The threshold of dimsim is 0.5. 

The system uses voice keywords F1 and actual term-weighted value (ATWV) to measure system performance. Among them, F1 reflects both the accuracy and the recall rate of the system. ATWV mainly evaluates the average TWV value of the system on each keyword, which reflects the system's detection effect on the keywords with different frequencies. 

The final submitted system is the ensemble of the three well-trained models.

\subsection{Semi-supervised Learning}
In order to solve the problem of domain mismatch between the training set and the validation set, the CN-Celeb dataset \cite{fan2020cn} is used for semi-supervised learning (SSL). We randomly select 560h CN-Celeb data for semi-supervised learning. The method we used are inspired by the noisy student training (NST). Formally, the selected unlabeled data and the well-trained teacher model are denoted as $U$ and $M_0$, respectively. The semi-supervised learning algorithm is summarized as follow:

(1) Set the initial teacher model $M=M_{0}$.

(2) Generate pseudo-label $M(U)$ via the teacher model $M$.

(3) Mix $M(U)$ and the VKW data to fine-tune the model $M_{0}$ and get a new model $M'$.

(4) Set $M=M'$ and repeat step (2) (3) until convergence.

\subsection{Experimental Results}
Table 1 shows the performance of the BBS-KWS system in three scenarios: long video (lgv), short video (stv), and live broadcast (liv). We can draw the following conclusions from the table. First, the language model, length normalization and syllable modeling units brought significant improvements. Second, compared to the chain baseline model, the F1 score of the BBS-KWS system is improved from (67.81\%, 65.65\%, 70.06\%) to (88.39\%, 89.73\%, 89.65\%) in the three scenarios respectively. And the ATWV score of the BBS-KWS system is improved from (51.71\%, 60.27\%, 56.44\%) to (84.95\%, 90.24\%, 88.95\%). Overall, the BBS-KWS achieves 31\% F1 relative increase, and 56\% ATWV increase. Among them, the best performance of the single system achieved 32\% F1 relative increase, and 52\% ATWV increase.

Furthermore, the semi-supervised learning uses the CN-Celeb data to increase the diversity of training data, thereby improving the performance of the model. This experiment is repeated four times, and the accuracy of pseudo-labels are improved through multiple rounds of iteration. As can be see from the Table 1, the SSL have brought a great improvement.

\section{Conclusions}

The BBS-KWS system exploits a hybrid CTC/Attention acoustic model, combined with a big backbone, syllable modeling units, and keyword biasing technology to improve the performance of the ASR module. In the KWS module, it uses the fuzzy matching and multi-stage matching methods, achieving promising performance. And we adopt the semi-supervised learning to further improve the robustness of the system. In the VKW task, the BBS-KWS system achieves significant gains over the baseline, which achieves 31\% F1 increase and 56\% ATWV increase.



\bibliographystyle{plain}

\bibliography{mybib}

\end{document}